%% file: main.tex
\documentclass[runningheads]{llncs}

\usepackage{accv}

\usepackage{accvabbrv}

\usepackage{graphicx}
\usepackage{booktabs}

\usepackage[accsupp]{axessibility}  

\usepackage[table]{xcolor}
\newcommand{\first}[1]{\cellcolor{yellow!40}\textbf{#1}}
\newcommand{\second}[1]{\cellcolor{cyan!20}\underline{#1}}
\newcommand{\third}[1]{\cellcolor{green!15}#1}

\usepackage{hyperref}

\usepackage{orcidlink}

\usepackage{float}

\begin{document}

\title{Split and Drive: Dual-Axis Disentanglement for Real-Time Gaussian Head Avatars}

\titlerunning{SpiD: Dual-Axis Disentanglement for Gaussian Head Avatars}

\author{MD Wahiduzzaman Khan\inst{1} \and
Mingshan Jia\inst{1} \and
Xiaolin Zhang\inst{2}\thanks{Corresponding author.} \and
En Yu\inst{1} \and
Kaska Musial-Gabrys\inst{1}}

\authorrunning{M.~W.~Khan et al.}

\institute{University of Technology Sydney, Australia\\
\email{arnobk511@gmail.com, mingshan.jia@uts.edu.au,}\\
\email{en.yu-1@uts.edu.au, musial.katarzyna@gmail.com}
\and
Shandong University of Science and Technology, China\\
\email{solli.zhang@gmail.com}}

\maketitle

\begin{figure}[H]
    \centering
    \includegraphics[width=\textwidth]{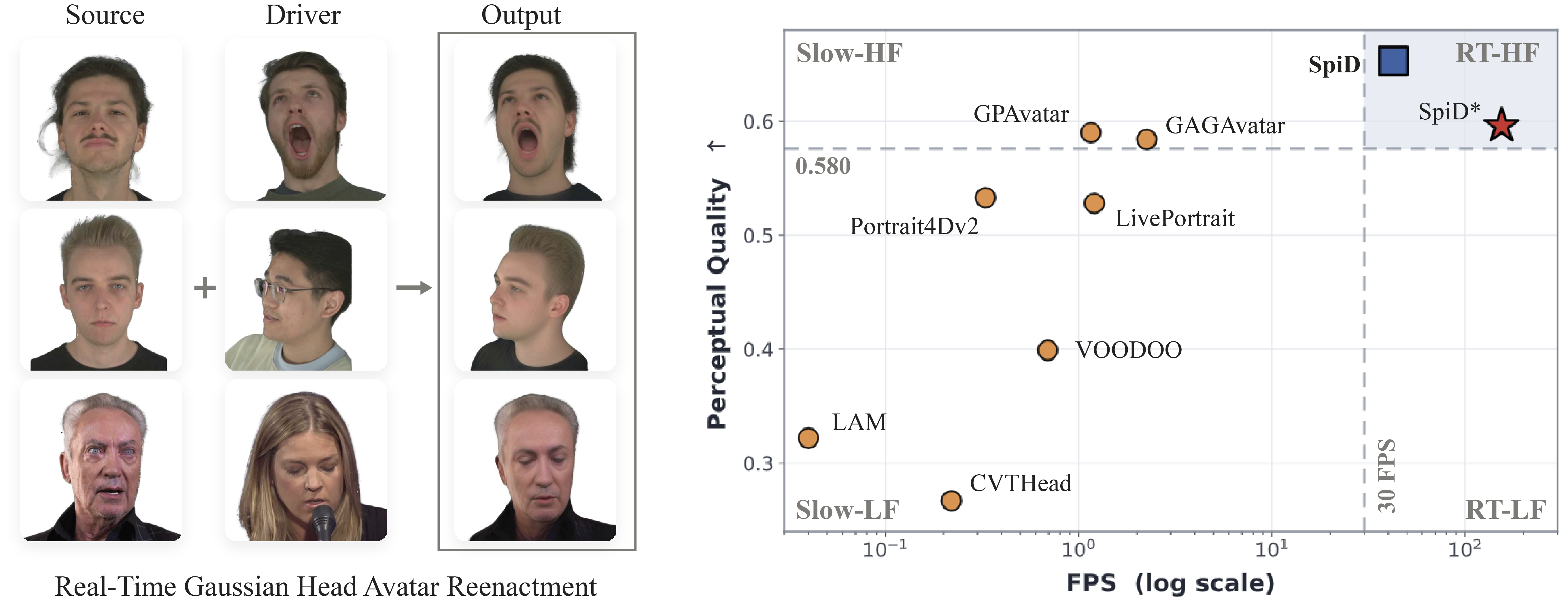}
    \caption{\textbf{SpiD} produces high-fidelity avatar animation from a single image across diverse identities and expressions (left). Among all compared methods, SpiD achieves the strongest quality-speed trade-off, placing in the real-time high-fidelity quadrant with the complete driving pipeline included (right). SpiD* denotes our speed variant without neural refinement.}
    \label{fig:teaser}
\end{figure}

\input{Sections/1_abstract}
\input{Sections/2_introduction}
\input{Sections/3_related_works}
\input{Sections/4_preliminaries}
\input{Sections/5_methodology}
\input{Sections/6_experiments}
\input{Sections/7_conclusion}

%
%
\bibliographystyle{splncs04}
\bibliography{main}

\end{document}

%% file: Sections/1_abstract.tex
\begin{abstract}
Creating photorealistic animatable head avatars from a single image remains a fundamental challenge in digital human synthesis. While recent 3D Gaussian Splatting methods have achieved promising results, they rely on external tracking pipelines whose latency is excluded from inference measurements. Furthermore, they adopt unified representations that entangle geometrically distinct facial regions, limiting both expressiveness and rendering fidelity. We propose \textbf{SpiD} (\textbf{Sp}l\textbf{i}t and \textbf{D}rive), a single-image Gaussian head avatar framework built on two disentanglement axes. The compute axis internalizes per-frame driving, eliminating external tracking dependency at inference. The feature axis decomposes the avatar into three specialized Gaussian branches, each modeling a geometrically distinct facial domain. Extensive experiments demonstrate consistently strong performance against state-of-the-art methods while achieving the fastest inference speed among all compared methods on a single GPU with the complete driving pipeline included.

\keywords{Face Reenactment \and Gaussian Splatting \and Generative Face Models}
\end{abstract}

%% file: Sections/2_introduction.tex
\section{Introduction}

Animatable 3D head avatars have become essential building blocks for immersive applications in virtual reality, telepresence, and digital entertainment. The advent of 3D Gaussian Splatting (3DGS)~\cite{kerbl20233d} has enabled real-time radiance field rendering with unprecedented efficiency, motivating a surge of single-image avatar methods. These approaches fall broadly into three families. The first leverages 2D generative models and implicit warping fields to animate portrait images~\cite{guo2024liveportrait, shi2026dex}. The second combines NeRF~\cite{mildenhall2021nerf} or 3DGS~\cite{kerbl20233d} with parametric face models such as FLAME~\cite{li2017learning} to achieve view-consistent animation~\cite{chu2024gpavatar, deng2024portrait4d, chu2024generalizable, kirschstein2026flexavatar, kirschstein2025avat3r}. The third employs diffusion-based generation~\cite{jiang2025instant} or large reconstruction models~\cite{kirschstein2025avat3r, he2025lam} to push the boundaries of visual quality and expressiveness. Despite this progress, achieving genuine real-time avatar animation remains elusive. Two fundamental limitations stand in the way. First, methods relying on parametric face models require external facial trackers at runtime. However, their cost is not reflected in standard benchmarks. Second, all three families handle the entire head with a single monolithic representation. This requires one network to cover every facial region regardless of its geometric properties.

Both limitations share a common architectural root: a failure to specialize. The first concerns how the driving pipeline is coupled to the avatar model, what we call the compute axis. The second concerns how the avatar representation is shared across facial regions, what we call the feature axis. Along the compute axis, the tracker is treated as a preprocessing step that runs before the avatar model is invoked, placing it entirely outside the model boundary. Since the driving pipeline and the avatar model remain structurally separate, end-to-end optimization of the full system is impossible, and genuine real-time deployment cannot be claimed. Along the feature axis, a single shared representation means every facial region competes for the same network capacity and feature space. Regions with fundamentally different geometric properties are handled by the same learned functions, and none receives the treatment its geometry actually demands. The result is a system that accounts for only part of the full pipeline and cannot fully serve every facial region.

To address this, we propose \textbf{SpiD} (\textbf{Sp}l\textbf{i}t and \textbf{D}rive), a real-time Gaussian head avatar framework built on dual-axis disentanglement. Along the compute axis, SpiD embeds a lightweight motion encoder directly within the avatar model and trains it end-to-end through the differentiable Gaussian renderer, so the encoder learns driving parameters optimized for rendering quality rather than geometric accuracy alone. Identity shape is estimated once and cached, requiring no external tracker at inference. This makes real-time performance an architectural guarantee of the system, not an artifact of selective benchmarking.

Along the feature axis, SpiD introduces three semantically non-overlapping Gaussian branches. The Dynamic Branch handles full head geometry, the Static Branch captures fine facial appearance, and the Mouth Interior Branch models oral dynamics that lie entirely outside the FLAME topology. SpiD is available in two variants: \textbf{SpiD} with the StyleUNet refiner for maximum quality, and \textbf{SpiD*} without the neural refiner for maximum speed, both achieving real-time performance with the complete driving pipeline included. As illustrated in Figure~\ref{fig:teaser}, SpiD achieves the strongest quality-speed trade-off among all compared methods, placing in the real-time high-fidelity quadrant with the complete driving pipeline included.

Our contributions are as follows:
\begin{itemize}
    \item \textbf{SpiD}, a real-time single-image 3DGS head avatar framework that unifies compute-axis and feature-axis disentanglement within a single end-to-end trainable system.
    \item \textbf{Compute-axis disentanglement} via a motion encoder embedded directly within the avatar model and supervised through the differentiable Gaussian renderer, eliminating external tracker dependency and reporting inference speed with the complete driving pipeline included.
    \item \textbf{Feature-axis disentanglement} via three geometrically non-overlapping Gaussian branches, each specialized for a distinct facial domain: full head geometry, facial appearance, and oral dynamics.
\end{itemize}

%% file: Sections/3_related_works.tex
\section{Related Work}

\noindent\textbf{3D Gaussian Splatting for Head Avatars.} The introduction of 3D Gaussian Splatting~\cite{kerbl20233d} has enabled real-time head avatar rendering at interactive frame rates. Early methods require per-subject optimization from multi-view recordings~\cite{kirschstein2023nersemble, grassal2022neural, hu2017avatar}, limiting generalization to unseen identities. A growing line of work therefore pursues generalizable feed-forward frameworks from one or a few images~\cite{chu2024generalizable, he2025lam, guo2025sega, kirschstein2026flexavatar, kirschstein2025avat3r}. Recent extensions of Gaussian representations to dynamic~\cite{luiten2024dynamic} and deformable~\cite{yang2024deformable} settings further motivate expression-driven deformation for head avatars. While these methods achieve strong 3D consistency, they rely on external facial trackers at inference time yet tracker latency is not reflected in reported inference speeds. Furthermore, they model geometrically heterogeneous facial regions within a single shared feature space, compromising rendering fidelity in regions that demand specialized treatment. SpiD addresses both limitations through dual-axis disentanglement.

\noindent\textbf{2D Portrait Animation.} A parallel line of work animates portrait images in 2D using keypoint-based~\cite{siarohin2019first} or neural rendering approaches~\cite{ren2021pirenderer, wang2021one}. GAN-based methods such as LivePortrait~\cite{guo2024liveportrait} achieve impressive inference speed through implicit keypoint warping, but lack true 3D geometric consistency under novel viewpoints and large pose variations. Diffusion-based approaches~\cite{shi2026dex, jiang2025instant} offer stronger expressiveness at the cost of 3D consistency and inference speed. In particular, Instant Expressive~\cite{jiang2025instant} distills motion priors from a 2D diffusion model into a 3D Gaussian representation, yet its performance is bounded by the quality of the synthetic training data generated by the teacher model. In contrast, SpiD operates on an explicit 3D representation throughout.

\noindent\textbf{Facial Tracking and Parametric Face Models.} Parametric face models~\cite{blanz2023morphable} provide a compact space for representing facial geometry. Prior work on 3D generative head models~\cite{chan2022efficient} further demonstrates the importance of geometry-aware representations for head synthesis. FLAME~\cite{li2017learning} has become a foundational parametric model for head avatar animation, providing a low-dimensional space for disentangling identity shape, pose, and expression. Prior methods for expressive face reconstruction~\cite{feng2021learning, danvevcek2022emoca} and NeRF-based head avatars~\cite{hong2022headnerf, guo2021ad} rely on such parametric models as geometric scaffolds. Most existing avatar methods treat FLAME tracking as an external preprocessing step whose latency is not captured by standard inference benchmarks~\cite{chu2024generalizable, he2025lam, kirschstein2026flexavatar}. SMIRK~\cite{retsinas2024smirk} advances expressive face reconstruction through an analysis-by-neural-synthesis training paradigm, producing more faithful reconstructions of extreme and asymmetric expressions. SpiD addresses this by fully internalizing the driving pipeline within the avatar model.

%% file: Sections/4_preliminaries.tex
\section{Preliminaries}

\noindent\textbf{3D Gaussian Splatting.} Following~\cite{kerbl20233d}, we represent the head avatar as a collection of anisotropic 3D Gaussian primitives. Each primitive is parameterized by its center position $\mu \in \mathbb{R}^3$, covariance matrix $\Sigma \in \mathbb{R}^{3\times3}$ decomposed as $\Sigma = RSS^\top R^\top$ where $R \in \mathbb{R}^{3\times3}$ is a rotation matrix and $S = \text{diag}(s_1, s_2, s_3)$ is a scaling matrix, opacity $o \in [0,1]$, and color feature $c \in \mathbb{R}^{32}$. The final rendered color of a pixel is computed via alpha compositing over depth-sorted Gaussians:
\begin{equation}
    \mathcal{C} = \sum_{i \in \mathcal{N}} c_i \alpha_i \prod_{j=1}^{i-1}(1 - \alpha_j)
\end{equation}
where $\alpha_i$ is obtained by evaluating the 2D projected Gaussian multiplied by its opacity $o_i$. Recent extensions have explored dynamic~\cite{luiten2024dynamic} and deformable~\cite{yang2024deformable} Gaussian representations for non-rigid scenes, motivating the use of expression-driven Gaussian deformation for head avatar animation.

\noindent\textbf{FLAME Parametric Face Model.} Parametric face models provide a compact, semantically meaningful space for representing facial geometry~\cite{blanz2023morphable}. We adopt FLAME~\cite{li2017learning} as our geometric prior, which generates a mesh of $N_v = 5023$ vertices as:
\begin{equation}
    \mathcal{M}(\beta, \psi, \theta) = W\left(\bar{T} + B_S(\beta) + B_E(\psi) + B_P(\theta),\ J(\beta),\ \theta,\ \mathcal{W}\right)
\end{equation}
where $\beta \in \mathbb{R}^{300}$ are shape coefficients, $\psi \in \mathbb{R}^{100}$ are expression coefficients, $\theta \in \mathbb{R}^{6}$ are pose parameters, $B_S$, $B_E$, $B_P$ are shape, expression, and pose blendshape functions respectively, $J(\beta)$ are the joint locations, and $\mathcal{W}$ are blend skinning weights. The resulting vertex set $\mathbf{V} = \mathcal{M}(\beta, \psi, \theta) \in \mathbb{R}^{N_v \times 3}$ serves as the geometric scaffold for our Dynamic and Mouth Interior branches.

%% file: Sections/5_methodology.tex
\section{Method}

\begin{figure*}[t!]
    \centering
    \includegraphics[width=\textwidth]{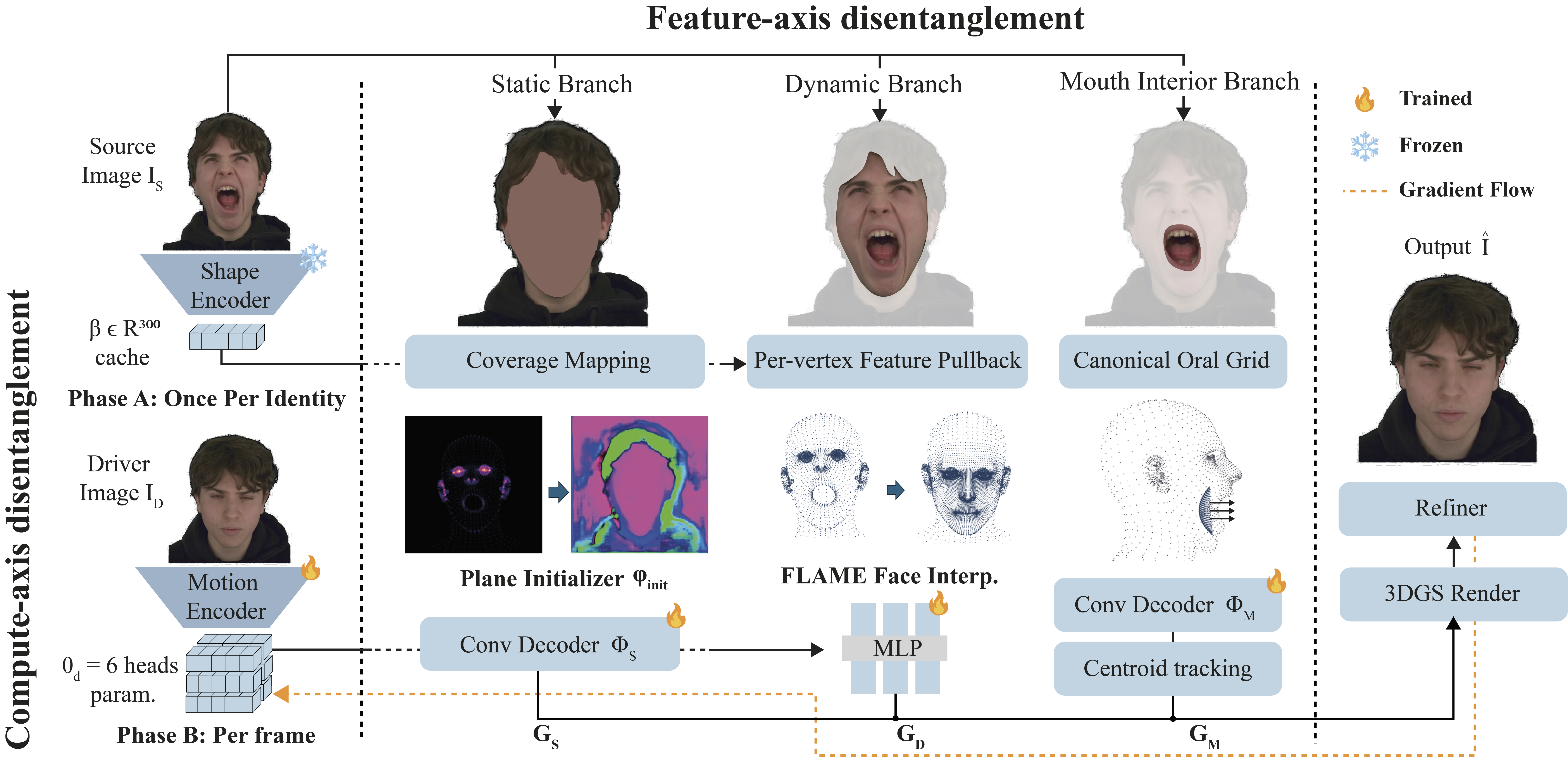}
    \caption{\textbf{Overview of SpiD.} Given a single source image and a driving frame, SpiD operates along two disentanglement axes. Along the compute axis, Phase A estimates identity shape once per subject via a frozen shape encoder, while Phase B predicts per-frame motion parameters $\Theta_d$ via a lightweight motion encoder. Along the feature axis, three specialized Gaussian branches process complementary spatial domains: the Static Branch models fine-grained facial appearance on a DINOv2-conditioned deformable image plane, the Dynamic Branch binds Gaussians to a densified FLAME mesh enriched with per-vertex DINOv2 features, and the Mouth Interior Branch operates in a canonical oral coordinate frame with live centroid tracking. The resulting Gaussians are rendered via 3DGS and refined by a neural refiner to produce the final output.}
    \label{fig:method}
\end{figure*}

Given a single source image $\mathcal{I}_s$ and a per-frame driving image $\mathcal{I}_d$, SpiD produces an avatar image $\hat{\mathcal{I}}$ that preserves the identity of $\mathcal{I}_s$ while transferring the expression and pose of $\mathcal{I}_d$. As illustrated in Figure~\ref{fig:method}, the pipeline operates along two disentanglement axes. Along the compute axis, a two-phase internalized driving module predicts all geometric driving signals directly from input images with no external dependencies (Section~\ref{sec:driving}). Along the feature axis, three geometrically specialized Gaussian branches decompose the avatar into complementary spatial domains (Section~\ref{sec:branches}). The resulting Gaussians are rendered via differentiable splatting~\cite{kerbl20233d} and refined by a neural upsampler (Section~\ref{sec:rendering}).

\subsection{Compute-Axis Disentanglement}
\label{sec:driving}

The central contribution of this axis is a driving encoder that is optimized not for geometric accuracy but for rendering quality. We achieve this by embedding the encoder directly within the avatar model and training it end-to-end through the differentiable Gaussian renderer, so that gradients from the photometric loss flow back through the rasterizer to the encoder weights. No prior single-image avatar method trains its driving parameters under this supervision signal. To ensure stability during early training, a geometric anchoring loss bootstraps the encoder with ground-truth tracked parameters, allowing the system to gradually shift from geometric to photometric supervision as training progresses.

\noindent\textbf{Phase A: Identity Shape Encoding.} Identity shape is a static property of the source subject that does not change across animation frames. We therefore compute it once and cache the result, contributing zero latency to per-frame inference. Given the source image $\mathcal{I}_s$, we extract a facial crop $\mathcal{I}_s^{224} \in \mathbb{R}^{224 \times 224 \times 3}$ via a differentiable FLAME-guided crop operation and predict identity-specific shape parameters using a frozen shape encoder $\mathcal{E}_\beta$:
\begin{equation}
    \beta = \mathcal{E}_\beta(\mathcal{I}_s^{224}) \in \mathbb{R}^{300}
\end{equation}

\noindent\textbf{Phase B: Per-Frame Motion Encoding.} Per-frame motion is predicted in real time from each driving image via six specialized linear heads, one per semantically distinct parameter group, each optimized jointly for visual output quality through the renderer. We extract a driving crop $\mathcal{I}_d^{224} \in \mathbb{R}^{224 \times 224 \times 3}$ and pass it through a trainable motion encoder $\mathcal{E}_m$~\cite{howard2019searching}, producing a feature vector $\mathbf{f}_d \in \mathbb{R}^{576}$:
\begin{equation}
    \mathbf{f}_d = \text{AvgPool}\left(\mathcal{E}_m(\mathcal{I}_d^{224})\right) \in \mathbb{R}^{576}
\end{equation}
Six lightweight linear heads predict the complete set of per-frame motion parameters $\Theta_d$:
\begin{equation}
    R = \phi_{\text{rot}}(\mathbf{f}_d) \in \mathbb{R}^{3\times3}, \quad \psi = \phi_{\text{exp}}(\mathbf{f}_d) \in \mathbb{R}^{100}
\end{equation}
\begin{equation}
    \theta_{\text{jaw}} = \phi_{\text{jaw}}(\mathbf{f}_d) \in \mathbb{R}^{3}, \quad \mathbf{t} = \phi_{\text{trans}}(\mathbf{f}_d) \in \mathbb{R}^{3}
\end{equation}
\begin{equation}
    \mathbf{e}_{\text{lid}} = \sigma(\phi_{\text{lid}}(\mathbf{f}_d)) \in \mathbb{R}^{2}, \quad \theta_{\text{eye}} = \phi_{\text{eye}}(\mathbf{f}_d) \in \mathbb{R}^{6}
\end{equation}
where $\sigma$ denotes the sigmoid activation and $R$ is obtained via 6D rotation representation~\cite{zhou2019continuity} for continuity. Eyelid closure is modeled explicitly via precomputed blendshape deltas $\Delta_L, \Delta_R \in \mathbb{R}^{N_v \times 3}$, capturing blink dynamics that expression coefficients alone cannot represent faithfully. The per-frame camera transformation is assembled as:
\begin{equation}
    T_d = \begin{bmatrix} R & \mathbf{t} \end{bmatrix} \in \mathbb{R}^{3\times4}
\end{equation}
and the driven FLAME vertices are computed as:
\begin{equation}
    \mathbf{V}_d = \mathcal{M}(\beta, \psi, \theta_{\text{jaw}}, \theta_{\text{eye}}) + e_{\text{lid},L} \cdot \Delta_L + e_{\text{lid},R} \cdot \Delta_R \in \mathbb{R}^{N_v \times 3}
\end{equation}
The pair $(\mathbf{V}_d, T_d)$ constitutes the complete per-frame geometric driving signal, produced entirely within the model with no external dependencies.

\subsection{Feature-Axis Disentanglement}
\label{sec:branches}

SpiD introduces three Gaussian branches: the Static Branch for facial appearance, the Dynamic Branch for full head geometry, and the Mouth Interior Branch for oral dynamics. Each branch operates in a dedicated coordinate frame, receives domain-aligned input features, and covers a spatially non-overlapping region of the head. A FLAME coverage map enforces this spatial exclusivity by explicitly preventing redundant Gaussian placement between branches.

\noindent\textbf{Static Branch.} The face surface contains the densest concentration of identity-relevant appearance detail, yet a fixed image-plane grid places Gaussians uniformly across the entire image, wasting capacity on regions already covered by the Dynamic Branch mesh. We therefore introduce a DINOv2-conditioned deformable plane initializer $\Phi_{\text{init}}$ that learns to reposition grid cells toward regions of high appearance complexity not covered by the mesh. A soft FLAME coverage map $\mathbf{C} \in [0,1]^{n \times n}$ computed from the source mesh signals which image regions the Dynamic Branch already models, and DINOv2 features~\cite{oquab2023dinov2} guide the initializer toward identity-discriminative surface regions. Both are concatenated as input:
\begin{equation}
    \mathbf{F}_{\text{in}} = [\mathcal{F}_{\text{DINO}}^{\text{skin}}(\mathcal{I}_s),\ \mathbf{C}] \in \mathbb{R}^{n \times n \times 257}
\end{equation}
The initializer predicts per-cell 2D offsets on top of a base uniform grid $\mathbf{U}_0 \in [-1,1]^{n \times n \times 2}$:
\begin{equation}
    \Delta\mathbf{U} = \Phi_{\text{init}}(\mathbf{F}_{\text{in}}) \cdot \delta_{\text{max}} \in \mathbb{R}^{n \times n \times 2}, \quad \delta_{\text{max}} = 0.05
\end{equation}
yielding deformed grid locations $\mathbf{U} = \mathbf{U}_0 + \Delta\mathbf{U}$. The deformed grid cells are lifted into 3D space by back-projecting each cell along its camera ray to a fixed focal plane:
\begin{equation}
    \mu_k^S = \mathbf{o}_s + d_s \cdot \mathbf{r}_k \in \mathbb{R}^3
\end{equation}
where $\mathbf{o}_s$ is the camera origin, $d_s$ the focal depth, and $\mathbf{r}_k$ the unit ray direction of grid cell $k$. A convolutional decoder $\Phi_S$ processes DINOv2~\cite{oquab2023dinov2} features resampled at deformed grid locations and produces the static Gaussian attributes, where $c_k^S \in \mathbb{R}^{32}$ denotes the color feature, $o_k^S \in [0,1]$ the opacity, $r_k^S \in \mathbb{R}^4$ the rotation quaternion, $s_k^S \in \mathbb{R}^3$ the scaling vector, and $p_k^S \in [0,1]$ a signed depth offset:
\begin{equation}
    \{c_k^S, o_k^S, r_k^S, s_k^S, p_k^S\} = \Phi_S\left([\mathbf{F}_{\text{DINO}},\ \mathbf{U}],\ \gamma(\mathbf{n})\right)
\end{equation}
where $\gamma(\mathbf{n}) \in \mathbb{R}^{27}$ is the harmonic encoding of the viewing direction. The predicted depth offset $p_k^S$ is applied along the plane normal $\mathbf{n}_s$:
\begin{equation}
    \mu_k^S \leftarrow \mu_k^S + (2p_k^S - 1) \cdot \mathbf{n}_s
\end{equation}
This signed offset enables the Static Branch to model subtle surface relief without departing from the image-plane coordinate frame.

\noindent\textbf{Dynamic Branch.} Hair, scalp, neck, and ears are geometrically rigid and expression-invariant, making them natural candidates for mesh-bound Gaussians that deform with the FLAME mesh per frame. The standard FLAME topology however provides only $N_v = 5023$ vertices, leaving these peripheral regions severely under-covered. We address this with edge-midpoint subdivision, computing a midpoint vertex for every edge in the FLAME triangle mesh:
\begin{equation}
    \mathbf{v}_{\text{mid}}^{(i,j)} = \frac{\mathbf{v}_i + \mathbf{v}_j}{2}
\end{equation}
yielding a densified vertex set $\tilde{\mathbf{V}} \in \mathbb{R}^{(N_v + |\mathcal{E}|) \times 3}$. Each densified vertex is then grounded in the source image by projecting it onto the source image plane and sampling the corresponding DINOv2~\cite{oquab2023dinov2} feature via bilinear interpolation:
\begin{equation}
    \mathbf{d}_k = \text{GridSample}\left(\mathcal{F}_{\text{DINO}}(\mathcal{I}_s),\ \mathbf{p}_k\right) \in \mathbb{R}^{256}
\end{equation}
where $\mathbf{p}_k \in [-1,1]^2$ is the NDC projection of vertex $\mathbf{v}_k$ onto the source image. This per-vertex feature pullback couples mesh geometry with source image appearance at the vertex level, providing identity-discriminative features that no global or UV-space encoding can offer. A lightweight MLP decoder $\Phi_D$ maps these enriched per-vertex features to Gaussian attributes:
\begin{equation}
    \{c_k^D, o_k^D, r_k^D, s_k^D\} = \Phi_D\left([\tilde{\mathbf{h}}_k,\ \mathbf{d}_k],\ \gamma(\mathbf{n})\right)
\end{equation}
Gaussian positions are set directly to driven densified mesh vertices $\mu_k^D = \tilde{\mathbf{V}}_d^{(k)}$, updated per-frame from $(\mathbf{V}_d, T_d)$.

\begin{figure*}[t!]
    \centering
    \includegraphics[width=\textwidth]{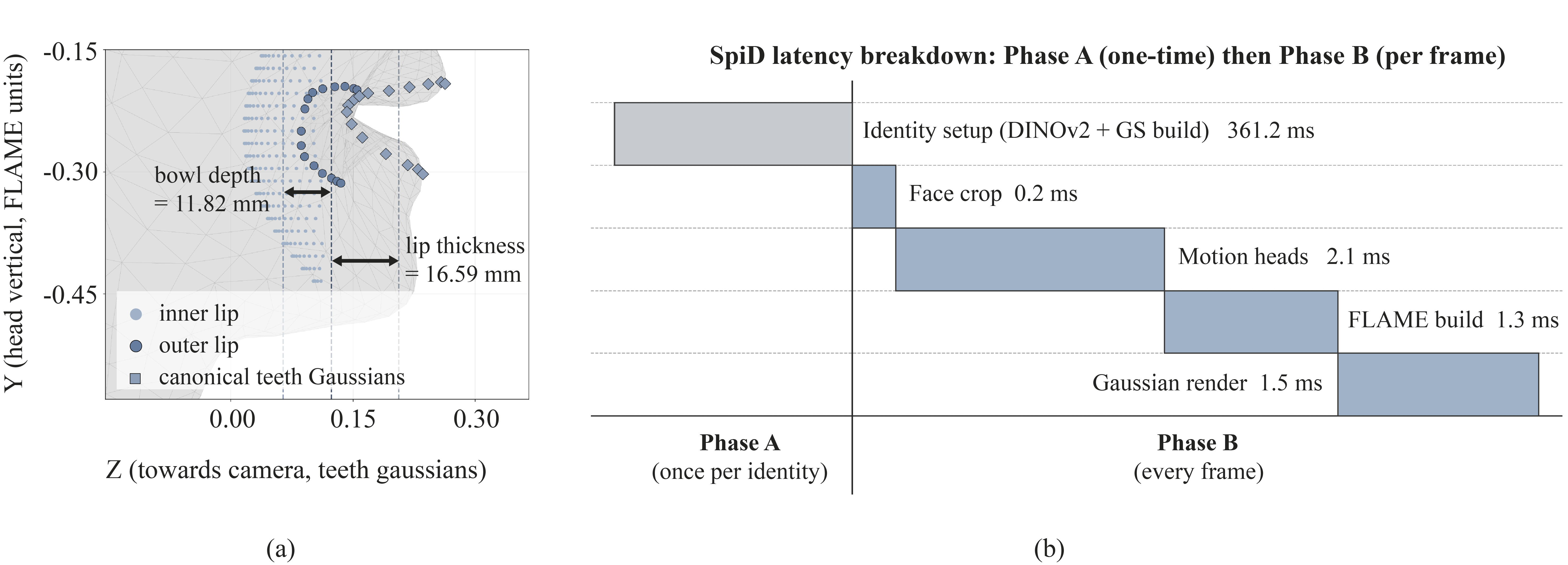}
    \caption{(a) Per-identity depth bound for the Mouth Interior Branch. Side view of the source FLAME geometry showing inner lip vertices (circles), outer lip vertices (diamonds), and canonical mouth Gaussians (dots). Two anatomical quantities define the admissible depth interval: $d_{\text{bowl}}$, the distance from the canonical interior sheet to the inner lip opening, and $d_{\text{lip}}$, the front-to-back thickness of the upper lip. Their sum with a small anatomical margin $\epsilon$ gives the per-identity cap, computed once from the source mesh. (b) SpiD latency breakdown showing Phase A (once per identity) and Phase B (per frame), demonstrating that the complete driving pipeline is fully internalized within the avatar model.}
    \label{fig:mouth_latency}
\end{figure*}

\noindent\textbf{Mouth Interior Branch.} The mouth interior lies entirely outside the FLAME surface topology and is therefore invisible to both the Static and Dynamic branches. Without explicit constraints, Gaussians in this region receive inconsistent supervision and drift in depth, producing smearing and flickering around open mouths. We introduce a dedicated branch that places a compact canonical sheet of Gaussians anchored at the inner lip ring centroid $\bar{\mathbf{l}}_s$, computed as the mean position of the inner lip ring vertices $\mathcal{V}_{\text{lip}}^{\text{inner}}$ on the source FLAME mesh. This centroid fixes the lateral, vertical, and depth position of the canonical sheet. A $K \times K$ grid of Gaussian positions $\mathbf{g}_{ij} \in \mathbb{R}^3$ is placed around $\bar{\mathbf{l}}_s$:
\begin{equation}
    \mathbf{g}_{ij} = \bar{\mathbf{l}}_s + r_{\text{lip}} \cdot [u_{ij},\ v_{ij},\ 0]^\top \in \mathbb{R}^3
\end{equation}
where $r_{\text{lip}}$ is the canonical lip radius and $(u_{ij}, v_{ij}) \in [-1,1]^2$ are canonical UV coordinates. A convolutional decoder $\Phi_M$ processes DINOv2~\cite{oquab2023dinov2} features sampled at each grid location concatenated with canonical UV coordinates, predicting per-Gaussian attributes and a learned local offset $\mathbf{q}_k^M$:
\begin{equation}
    \{c_k^M, o_k^M, r_k^M, s_k^M, \mathbf{q}_k^M\} = \Phi_M\left([\mathbf{F}_{\text{DINO}}^{\text{oral}},\ \mathbf{UV}^{\text{oral}}],\ \gamma(\mathbf{n})\right)
\end{equation}
At each animation frame, the entire grid translates rigidly from the source to the driven lip centroid, following mouth dynamics without any per-Gaussian deformation field:
\begin{equation}
    \mu_k^M = \mathbf{g}_{ij} - \bar{\mathbf{l}}_s + \bar{\mathbf{l}}_d + \mathbf{q}_k^M \in \mathbb{R}^3
\end{equation}
where $\bar{\mathbf{l}}_d$ is the driven lip centroid from the driven FLAME vertices. To prevent depth drift, we derive a per-identity depth cap from two anatomical quantities on the source mesh (Figure~\ref{fig:mouth_latency}a): the bowl depth $d_{\text{bowl}}$ and half the upper lip thickness $d_{\text{lip}}$:
\begin{equation}
    \text{cap} = d_{\text{bowl}} + \frac{1}{2} d_{\text{lip}} + \epsilon
\end{equation}
The depth component of $\mathbf{q}_k^M$ is clamped to this cap. For a typical subject $d_{\text{bowl}} \approx 11.82$ mm and $d_{\text{lip}} \approx 16.59$ mm.

\subsection{Rendering and Training}
\label{sec:rendering}

\noindent\textbf{Rendering and Neural Refinement.} The complete set of Gaussian primitives is obtained by merging outputs from all three branches:
\begin{equation}
    \mathcal{G} = \mathcal{G}_D \cup \mathcal{G}_S \cup \mathcal{G}_M
\end{equation}
These primitives are rendered to a 32-channel feature image $\hat{\mathcal{I}}_{\text{raw}} \in \mathbb{R}^{H \times W \times 32}$ via the differentiable tile-based rasterizer of~\cite{kerbl20233d}, with the first three channels forming the coarse RGB render $\hat{\mathcal{I}}_c \in \mathbb{R}^{H \times W \times 3}$. A StyleUNet neural upsampler $\Phi_{\text{SR}}$ produces the final output:
\begin{equation}
    \hat{\mathcal{I}} = \Phi_{\text{SR}}(\hat{\mathcal{I}}_{\text{raw}}) \in \mathbb{R}^{H \times W \times 3}
\end{equation}
By operating in the shared 32-dimensional feature space rather than on raw RGB values, the upsampler achieves coherent fusion of all three branch contributions without boundary artifacts at their spatial boundaries.

\noindent\textbf{Training Objectives.} We apply an L1 reconstruction loss at both coarse and super-resolved levels, on both the full image and a cropped face region for higher-resolution supervision on identity-critical details:
\begin{equation}
    \mathcal{L}_{\text{img}} = \|\hat{\mathcal{I}}_c - \mathcal{I}_t\|_1 + \|\hat{\mathcal{I}} - \mathcal{I}_t\|_1 + \|\text{Crop}(\hat{\mathcal{I}}_c) - \text{Crop}(\mathcal{I}_t)\|_1 + \|\text{Crop}(\hat{\mathcal{I}}) - \text{Crop}(\mathcal{I}_t)\|_1
\end{equation}
A pretrained face perceptual network $\Phi_{\text{perc}}$~\cite{zhang2018unreasonable} is applied to the cropped face region to supervise high-frequency appearance:
\begin{equation}
    \mathcal{L}_{\text{perc}} = \|\Phi_{\text{perc}}(\text{Crop}(\hat{\mathcal{I}}_c)) - \Phi_{\text{perc}}(\text{Crop}(\mathcal{I}_t))\|_1 + \|\Phi_{\text{perc}}(\text{Crop}(\hat{\mathcal{I}})) - \Phi_{\text{perc}}(\text{Crop}(\mathcal{I}_t))\|_1
\end{equation}
Geometric anchor losses against ground-truth tracked FLAME parameters keep the internalized driving encoder well-grounded throughout training:
\begin{equation}
    \mathcal{L}_{\text{aux}} = \lambda_v \mathcal{L}_{\text{verts}} + \lambda_r \mathcal{L}_{\text{rot}} + \lambda_t \mathcal{L}_{\text{trans}} + \lambda_e \mathcal{L}_{\text{eye}}
\end{equation}
A differentiable landmark loss using a frozen face alignment network~\cite{bulat2017far} enforces cross-pose geometric consistency:
\begin{equation}
    \mathcal{L}_{\text{lmk}} = \left\|\frac{\hat{\mathbf{l}} - \mathbf{l}^{\text{gt}}}{64}\right\|_2^2
\end{equation}
The complete training objective is:
\begin{equation}
    \mathcal{L} = \lambda_1 \mathcal{L}_{\text{img}} + \lambda_2 \mathcal{L}_{\text{perc}} + \lambda_3 \mathcal{L}_{\text{aux}} + \lambda_4 \mathcal{L}_{\text{lmk}}
\end{equation}

%% file: Sections/6_experiments.tex
\section{Experiments}

\subsection{Implementation Details}
We implement SpiD in PyTorch and train on a combination of the VFHQ~\cite{xie2022vfhq} and NeRSemble~\cite{kirschstein2023nersemble} datasets, sampling paired source and driving frames from the same video clip. All images are resized to $512 \times 512$. The DINOv2 ViT-B/14 backbone~\cite{oquab2023dinov2} is frozen throughout training. We train using the Adam optimizer with a learning rate of $1 \times 10^{-4}$ and batch size 8 on a single NVIDIA A100 GPU for 500k steps. The local plane grid size is $n = 296$ and the mouth interior grid is $K \times K = 32 \times 32$. Loss weights are set to $\lambda_1 = 1.0$, $\lambda_2 = 1.0$, $\lambda_3 = 1.0$, $\lambda_4 = 1.0$, with internal component weights for $\mathcal{L}_{\text{aux}}$ set to 100, 10, 5, and 10 for vertex, rotation, translation, and eye supervision respectively. We report two variants: \textbf{SpiD} with the StyleUNet refiner for maximum quality, and \textbf{SpiD*} without the StyleUNet refiner for maximum speed. All reported inference speeds are measured on a single NVIDIA A100 GPU and include the complete driving pipeline with no components excluded.

\subsection{Baselines}
We compare against the following state-of-the-art methods. GAGAvatar~\cite{chu2024generalizable} reconstructs 3D Gaussians from a single image via dual-lifting with a FLAME-driven expression branch. GPAvatar~\cite{chu2024gpavatar} uses a point-based expression field with a tri-plane canonical representation. LAM~\cite{he2025lam} scales single-image avatar generation with a large-scale model trained on both 2D and 3D data. Portrait4D-v2~\cite{deng2024portrait4d} learns dynamic expression tri-planes from pseudo multi-view videos. LivePortrait~\cite{guo2024liveportrait} is a GAN-based 2D portrait animation method using implicit keypoint representations~\cite{siarohin2019first}. VOODOO-XP is a volumetric disentanglement method for cross-person expression transfer. CVTHead is a point-based neural rendering approach for head avatar reconstruction. All baselines are evaluated using their official pretrained models and inference pipelines.

\subsection{Evaluation Metrics}
We adopt a comprehensive set of metrics covering identity preservation, expression accuracy, perceptual quality, and pixel-level fidelity. For identity we use ArcFace~\cite{deng2019arcface} cosine similarity with the source (ArcF(src)$\uparrow$) and with the driver (ArcF(drv)$\downarrow$), where a low ArcF(drv) indicates the method does not leak driver identity into the output. For expression accuracy we use Average Keypoint Distance (AKD$\downarrow$), Average Expression Distance (AED$\downarrow$), and Average Pose Distance (APD$\downarrow$)~\cite{siarohin2019first}. For perceptual quality we use LPIPS~\cite{zhang2018unreasonable}, MUSIQ~\cite{ke2021musiq}, and TopIQ~\cite{chen2024topiq}. For pixel-level fidelity we use PSNR$\uparrow$, SSIM~\cite{wang2004image}$\uparrow$, MS-SSIM~\cite{wang2003multiscale}$\uparrow$, and L1$\downarrow$.

\subsection{Comparison with State of the Art}

\begin{figure*}[t!]
    \centering
    \includegraphics[width=0.9\textwidth]{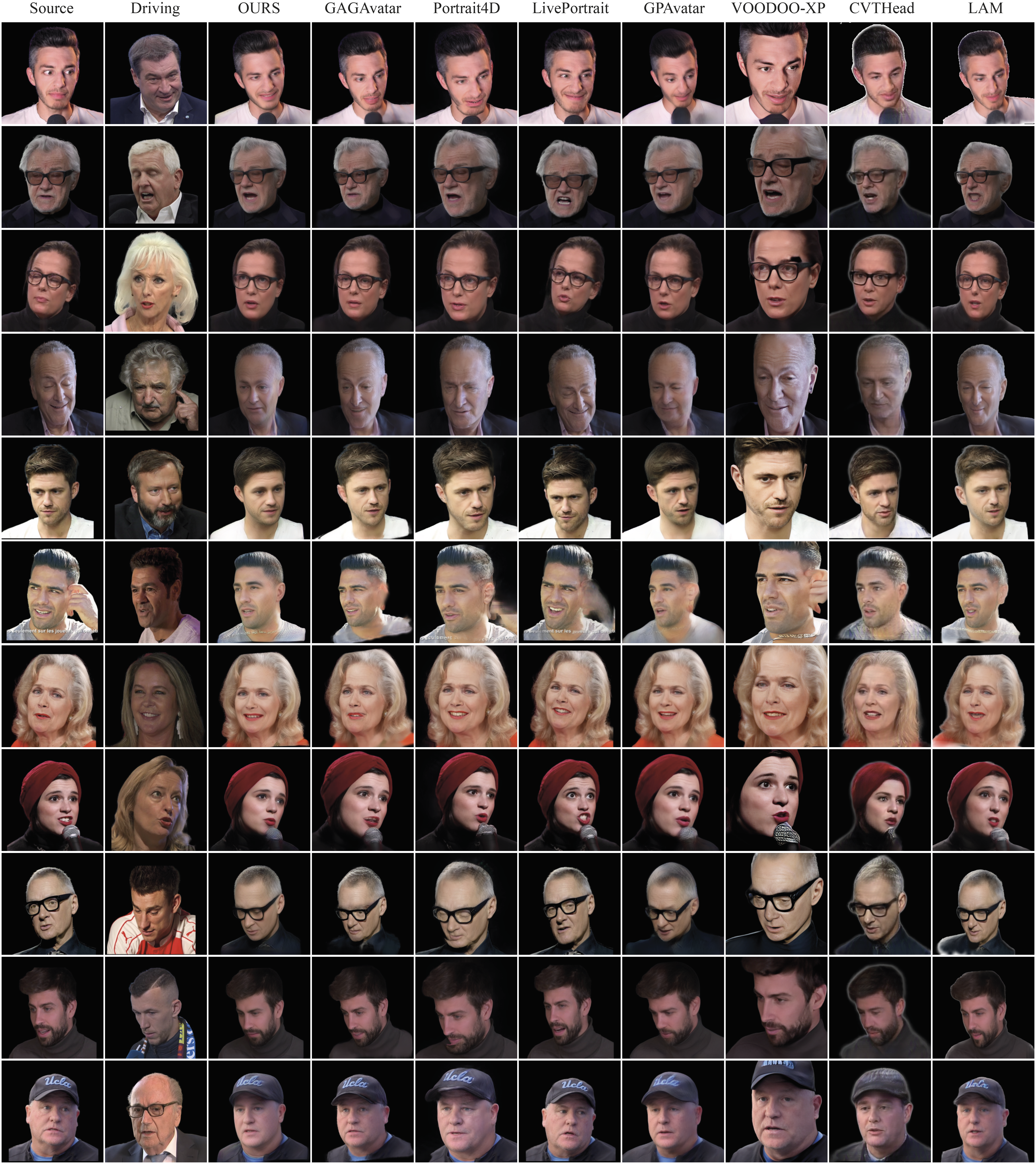}
    \caption{Qualitative comparison against state-of-the-art methods on cross-identity reenactment. SpiD produces more faithful expression transfer and sharper appearance reconstruction across diverse identities.}
    \label{fig:comparison}
\end{figure*}

\noindent\textbf{Cross-Identity Reenactment.} Table~\ref{tab:cross} reports cross-identity reenactment results following the evaluation protocol of prior methods~\cite{chu2024generalizable, he2025lam}, sampling 100 source identities from VFHQ~\cite{xie2022vfhq} driven by diverse expressions and pose variations. SpiD achieves first place in 7 out of 10 metrics including AKD, APD, LPIPS, PSNR, SSIM, MS-SSIM, and L1, while SpiD* achieves second place on all pixel-level fidelity metrics. Portrait4D-v2 and VOODOO-XP achieve higher no-reference quality scores (MUSIQ, TopIQ) by producing visually smooth outputs at the cost of expression accuracy. Figure~\ref{fig:comparison} shows qualitative comparisons against all baselines. SpiD produces sharper facial details and more accurate expression transfer, particularly in the mouth region where the dedicated Mouth Interior Branch provides explicit coverage of a region not explicitly modeled by any compared method.

\begin{table*}[t!]
\centering
\caption{Cross-identity reenactment on VFHQ~\cite{xie2022vfhq}. \colorbox{yellow!40}{\textbf{1st}}, \colorbox{cyan!20}{\underline{2nd}}, \colorbox{green!15}{3rd}.}
\label{tab:cross}
\resizebox{\textwidth}{!}{
\begin{tabular}{lcccccccccc}
\toprule
Method & AKD$\downarrow$ & AED$\downarrow$ & APD$\downarrow$ & LPIPS$\downarrow$ & PSNR$\uparrow$ & SSIM$\uparrow$ & MS-SSIM$\uparrow$ & L1$\downarrow$ & MUSIQ$\uparrow$ & TopIQ$\uparrow$ \\
\midrule
GAGAvatar~\cite{chu2024generalizable}   & 10.28          & 9.90           & 11.42          & 0.413          & \third{13.44}  & 0.567          & 0.525          & \third{0.130}  & 64.0           & 0.511          \\
GPAvatar~\cite{chu2024gpavatar}          & \third{7.77}   & \third{6.72}   & \third{10.93}  & \third{0.411}  & 13.38          & \third{0.569}  & \third{0.531}  & 0.133          & 62.5           & 0.476          \\
Portrait4D-v2~\cite{deng2024portrait4d} & 18.26          & 17.53          & 20.46          & 0.466          & 12.68          & 0.456          & 0.455          & 0.152          & \first{67.5}   & \second{0.558} \\
LivePortrait~\cite{guo2024liveportrait} & 46.37          & 49.16          & 38.01          & 0.456          & 13.02          & 0.541          & 0.471          & 0.138          & 56.3           & 0.400          \\
VOODOO-XP                               & 52.54          & 52.70          & 52.06          & 0.604          & 10.90          & 0.392          & 0.349          & 0.199          & \third{65.2}   & \first{0.589}  \\
CVTHead                                 & 19.75          & 20.16          & 18.52          & 0.713          & 3.27           & 0.174          & 0.174          & 0.566          & 36.2           & 0.281          \\
LAM~\cite{he2025lam}                    & \second{7.25}  & \first{6.08}   & \second{10.75} & 0.665          & 2.92           & 0.179          & 0.167          & 0.599          & 60.4           & 0.474          \\
\midrule
SpiD* (ours)                            & 9.98           & 9.13           & 12.53          & \second{0.393} & \second{13.73} & \second{0.589} & \second{0.536} & \second{0.122} & 60.8           & 0.456          \\
\textbf{SpiD (ours)}                    & \first{7.05}   & \second{6.45}  & \first{8.85}   & \first{0.352}  & \first{14.50}  & \first{0.627}  & \first{0.593}  & \first{0.111}  & \second{66.8}  & \third{0.527}  \\
\bottomrule
\end{tabular}}
\end{table*}

\noindent\textbf{Self-Reenactment.} Table~\ref{tab:self} reports self-reenactment results on VFHQ~\cite{xie2022vfhq}. SpiD achieves first place in 5 out of 10 metrics including PSNR, SSIM, MS-SSIM, LPIPS, and L1, demonstrating the strongest pixel-level and perceptual reconstruction quality among all compared methods. GAGAvatar and GPAvatar achieve the strongest expression accuracy and identity preservation, benefiting from per-identity optimization and a dense point-based expression field respectively. SpiD remains competitive on these metrics while leading decisively on reconstruction fidelity.

\begin{table}[t!]
\centering
\caption{Self-reenactment on VFHQ~\cite{xie2022vfhq}. \colorbox{yellow!40}{\textbf{1st}}, \colorbox{cyan!20}{\underline{2nd}}, \colorbox{green!15}{3rd}.}
\label{tab:self}
\resizebox{\columnwidth}{!}{
\begin{tabular}{lcccccccccc}
\toprule
Method & PSNR$\uparrow$ & SSIM$\uparrow$ & MS-SSIM$\uparrow$ & LPIPS$\downarrow$ & L1$\downarrow$ & MUSIQ$\uparrow$ & ArcF$\uparrow$ & AKD$\downarrow$ & AED$\downarrow$ & APD$\downarrow$ \\
\midrule
GAGAvatar~\cite{chu2024generalizable}   & \second{23.08} & \second{0.804} & \second{0.843} & \second{0.133} & \second{0.028} & \second{66.2}  & \third{0.890}  & \second{4.33}  & \second{4.03}  & \first{5.24}   \\
GPAvatar~\cite{chu2024gpavatar}          & \third{22.30}  & \third{0.794}  & \third{0.832}  & \third{0.157}  & \third{0.031}  & 64.0           & \first{0.905}  & \first{3.90}   & \first{3.39}   & \second{5.45}  \\
Portrait4D-v2~\cite{deng2024portrait4d} & 16.86          & 0.589          & 0.577          & 0.318          & 0.077          & \first{66.4}   & \second{0.898} & 19.51          & 19.36          & 19.97          \\
LivePortrait~\cite{guo2024liveportrait} & 18.95          & 0.700          & 0.663          & 0.251          & 0.056          & 56.4           & 0.846          & 24.84          & 25.52          & 22.79          \\
VOODOO-XP                               & 12.62          & 0.462          & 0.409          & 0.543          & 0.154          & 63.9           & 0.820          & 53.59          & 52.27          & 57.56          \\
\midrule
SpiD* (ours)                            & 21.30          & 0.756          & 0.765          & 0.180          & 0.039          & 62.1           & 0.865          & 8.79           & 8.37           & 10.06          \\
\textbf{SpiD (ours)}                    & \first{24.31}  & \first{0.812}  & \first{0.857}  & \first{0.125}  & \first{0.026}  & \third{66.0}   & 0.857          & \third{5.00}   & \third{4.83}   & \third{5.50}   \\
\bottomrule
\end{tabular}}
\end{table}

\subsection{Inference Speed Analysis}
SpiD fully internalizes the per-frame driving pipeline within the avatar model, ensuring that all reported inference speeds reflect the complete end-to-end system. As detailed in Figure~\ref{fig:mouth_latency}b, Phase B comprises face crop (0.2 ms), motion encoder and linear heads (2.1 ms), FLAME mesh build (1.3 ms), and Gaussian render (1.5 ms), totalling 6.1 ms per frame on a single NVIDIA A100 GPU. Phase A runs once per identity and is amortized across all animation frames at no per-frame cost. SpiD achieves 43 FPS and SpiD* achieves 154 FPS, both with the complete driving pipeline included.

\subsection{Ablation Study}
\begin{figure*}[t!]
    \centering
    \includegraphics[width=\textwidth]{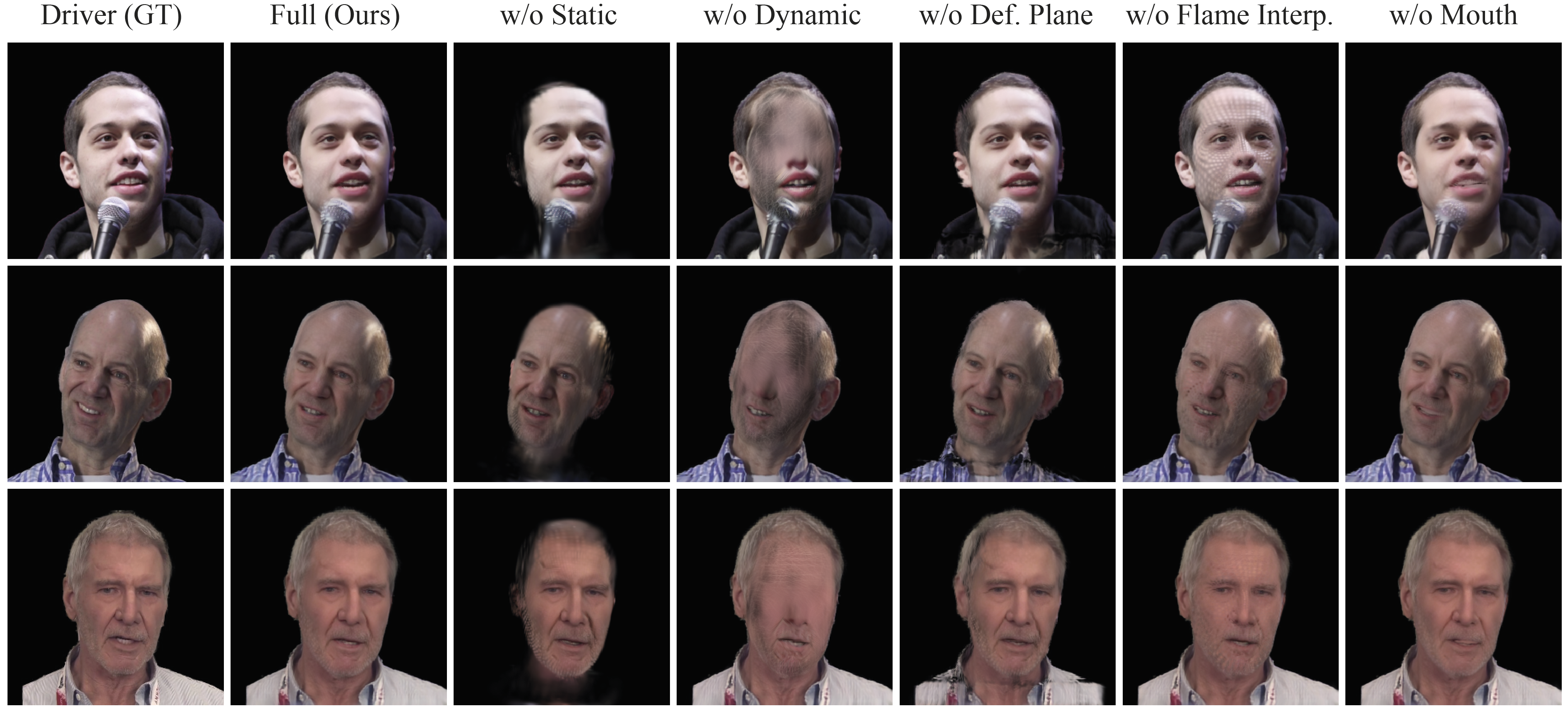}
    \caption{Qualitative ablation results. Removing the Static Branch washes out fine facial detail. Removing the Dynamic Branch causes identity collapse. Removing FLAME mesh interpolation produces visible mesh artifacts on the face surface. Replacing the Deformable Plane with a fixed grid reduces local appearance fidelity. The Mouth Interior Branch contribution is most evident in open-mouth cross-reenactment scenarios.}
    \label{fig:ablation}
\end{figure*}

We conduct ablation studies on VFHQ~\cite{xie2022vfhq} self-reenactment to validate each design choice in SpiD. Table~\ref{tab:ablation} presents results where each ablated component degrades on a distinct and complementary metric, directly validating our feature-axis disentanglement design. Qualitative results in Figure~\ref{fig:ablation} further illustrate the distinct failure mode of each ablated component. Removing the Dynamic Branch causes catastrophic identity collapse, with CSIM dropping from 0.829 to 0.281 and severe expression degradation, confirming that the Dynamic Branch is the primary carrier of identity and full-head geometry. Removing the Static Branch causes the most severe perceptual degradation with LPIPS increasing from 0.178 to 0.375, demonstrating that the Static Branch is the primary contributor to fine-grained facial appearance. Removing FLAME mesh interpolation produces visible mesh wireframe artifacts on the face surface with significant identity drift (CSIM $0.829 \rightarrow 0.695$), confirming that edge-midpoint subdivision is critical for seamless geometric coverage. Replacing the DINOv2-conditioned deformable plane with a fixed grid degrades perceptual quality (LPIPS $0.178 \rightarrow 0.206$), confirming that content-aware plane deformation improves local appearance capture.

\begin{table}[t!]
\centering
\caption{Ablation study on VFHQ~\cite{xie2022vfhq} self-reenactment. Each branch degrades on a distinct and complementary axis, validating the feature-axis disentanglement design.}
\label{tab:ablation}
\resizebox{\columnwidth}{!}{
\begin{tabular}{lccccccl}
\toprule
Variant & PSNR$\uparrow$ & SSIM$\uparrow$ & LPIPS$\downarrow$ & CSIM$\uparrow$ & AED$\downarrow$ & AKD$\downarrow$ & Primary effect \\
\midrule
\textbf{SpiD* (Full)} & \textbf{20.82} & \textbf{0.760} & \textbf{0.178} & \textbf{0.829} & \textbf{6.75} & \textbf{6.84} & -- \\
\midrule
w/o Dynamic Branch   & 19.52 & 0.737 & 0.230 & \cellcolor{red!15}0.281    & 16.46 & 14.95 & Identity collapse \\
w/o Static Branch    & 15.87 & 0.677 & \cellcolor{red!15}0.375    & 0.777 & 6.98  & 7.27  & Appearance loss \\
w/o FLAME Interp.    & 20.35 & 0.739 & 0.206 & \cellcolor{orange!15}0.695 & 7.24  & 7.33  & Identity drift \\
w/o Deformable Plane & 20.38 & 0.753 & \cellcolor{orange!15}0.206 & 0.820 & 6.76  & 6.87  & Fidelity drop \\
\bottomrule
\end{tabular}}
\end{table}

%% file: Sections/7_conclusion.tex
\section{Conclusion}

We presented \textbf{SpiD}, a real-time single-image Gaussian head avatar framework built on dual-axis disentanglement. Along the compute axis, we fully internalized the driving pipeline within the avatar model, making real-time performance an architectural guarantee rather than a reporting choice. Along the feature axis, we introduced three geometrically specialized Gaussian branches, each designed to model exactly the facial domain its geometry demands. Extensive experiments demonstrate consistently strong performance against state-of-the-art methods while achieving the fastest inference speed with the complete driving pipeline included.
Identity-specific shape estimation currently relies on a single source image crop, which may be insufficient for subjects with significant self-occlusions or extreme head poses. The framework also bakes illumination from the source image into the avatar representation, limiting applicability under varying lighting conditions. We believe SpiD establishes a new standard for real-time benchmarking in head avatar synthesis, and we hope it encourages future work to report inference speed with the complete driving pipeline included. Future directions include dynamic illumination disentanglement, richer oral geometry modeling, and extension to audio-driven avatar animation.